\documentclass{article}

\usepackage{arxiv}

\usepackage[utf8]{inputenc} % allow utf-8 input
\usepackage[T1]{fontenc}    % use 8-bit T1 fonts
\usepackage{hyperref}       % hyperlinks
\usepackage{url}            % simple URL typesetting
\usepackage{booktabs}       % professional-quality tables
\usepackage{amsmath,amssymb,amsfonts}
\usepackage{nicefrac}       % compact symbols for 1/2, etc.
\usepackage{microtype}      % microtypography
\usepackage{cleveref}       % smart cross-referencing
\usepackage{lipsum}         % Can be removed after putting your text content
\usepackage{graphicx}
\usepackage{natbib}

\usepackage{textcomp}
\usepackage{xcolor}
\usepackage{subcaption}
\usepackage{multirow}
\usepackage[linesnumbered,ruled,vlined]{algorithm2e}

\title{Masked Multi-Step Multivariate Time Series Forecasting with Future Information}

\date{}

\author{Yiwei Fu\thanks{Corresponding author.} \\
	Machine Learning\\
	GE Research\\
	Niskayuna, NY 12309 \\
	\texttt{yiwei.fu@ge.com} \\
	\And
	Honggang Wang\thanks{Now at Upstart Power, Inc.} \\
	Power Systems\\
	GE Research\\
	Niskayuna, NY 12309 \\
	\texttt{hgwang2010@gmail.com} \\
	\And
	Nurali Virani\thanks{Project lead.} \\
	Machine Learning\\
	GE Research\\
	Niskayuna, NY 12309 \\
	\texttt{nurali.virani@ge.com} \\
}

\renewcommand{\shorttitle}{Masked Multi-Step Multivariate Time Series Forecasting with Future Information}

\begin{document}
\maketitle

\begin{abstract}
In this paper, we introduce Masked Multi-Step Multivariate Forecasting (MMMF), a novel and general self-supervised learning framework for time series forecasting with known future information.
In many real-world forecasting scenarios, some future information is known, e.g., the weather information when making a short-to-mid-term electricity demand forecast, or the oil price forecasts when making an airplane departure forecast.
Existing machine learning forecasting frameworks can be categorized into (1) sample-based approaches where each forecast is made independently, and (2) time series regression approaches where the future information is not fully incorporated.
To overcome the limitations of existing approaches, we propose MMMF, a framework to train any neural network model capable of generating a sequence of outputs, that combines both the temporal information from the past and the known information about the future to make better predictions.
Experiments are performed on two real-world datasets for (1) mid-term electricity demand forecasting, and (2) two-month ahead flight departures forecasting.
They show that the proposed MMMF framework outperforms not only sample-based methods but also existing time series forecasting models with the exact same base models.
Furthermore, once a neural network model is trained with MMMF, its inference speed is similar to that of the same model trained with traditional regression formulations, thus making MMMF a better alternative to existing regression-trained time series forecasting models if there is some available future information.
\end{abstract}

% keywords can be removed

\section{Introduction}~\label{sec:intro}
Time series forecasting that requires multi-step predictions has become an important part of many real-world applications in areas such as electricity demand modeling, air traffic volume prediction, stock prices forecasting, and crop yields estimation~\citep{cheng2006multistep}.
Specifically, there is often some future information available, such as the weather information for short-to-mid-term electricity demand modeling, and jet fuel price for air traffic volume prediction, which is not fully leveraged in the existing forecasting frameworks.
In recent years, modern machine learning techniques that enable the purely data-driven way of learning temporal dynamics have become more popular~\citep{ahmed2010empirical}.
In particular, since deep learning has gained significant grounds in many application domains, it has also been widely used in time series modeling by learning complex data representations without the need for manual feature engineering~\citep{lim2021time}.

For deep learning to work with time series forecasting, we need to first choose the appropriate basic building blocks, i.e., neural network (NN) models that are suitable for them. Three kinds of NN models are often used: Recurrent Neural Networks (RNNs), Convolutional Neural Networks (CNNs), and attention-based methods. 
RNNs, especially Long Short-term Memory (LSTM) networks~\citep{hochreiter_long_1997} which address the vanishing and exploding gradients problems in vanilla RNNs, are the classical method for modeling sequential data by keeping an internal memory state and updating them recursively with each new observation.
CNNs are originally designed for image-related tasks~\citep{lecun1989backpropagation}, which have been adapted to handle time series data by using causal convolution layers~\citep{bai_empirical_2018}.
Attention-based methods, such as Transformers~\citep{vaswani_attention_2017}, are designed to improve the learning of long-term dependencies and have shown state-of-the-art performance in many natural language processing tasks~\citep{devlin_bert_2019}.

With these base models as basic building blocks, the multi-step time series forecasting problem can be categorized into two kinds of approaches: recursive methods and direct methods~\citep{lim2021time}.
Recursive methods typically use an autoregressive approach (one-step-ahead prediction) and produce multi-step forecasts by recursively feeding predictions as inputs into the future time steps.
However, there is often some error at each step, so the recursive structure tends to accumulate large errors over long forecasting horizons.
Direct methods, on the other hand, directly map all available inputs to multi-step forecasts and typically use a sequence-to-sequence (seq2seq) structure.
The disadvantage of this method is that it is harder to train, especially when the forecast horizon is large~\citep{kline2004methods}.
It should be noted that we can also treat the multi-step forecasting problem as predicting each independent sample instead of a time series, if there are some predictor variables for each time step, and use a fully connected NN to achieve that.

These existing deep learning time series methods, however, do not incorporate known future information directly during training.
Recursive methods could use the future information when making iterative forecasts, but they are trained on 1-step predictions only.
Partially inspired by the success of masked language models such as BERT~\citep{devlin_bert_2019}, and to address the gap in incorporating known future information in multi-step forecasting, we propose Masked Multi-Step Multivariate Forecasting (MMMF).
MMMF is not a model, but a general self-supervised learning task for training all aforementioned time series models (including RNNs, CNNs, and attention-based models) to make multi-step forecasts with known future information.
MMMF is a flexible learning framework that improves upon existing methods by taking into account both recent history and known future information.

The contributions of this paper are as follows:
\begin{itemize}
	\item We propose MMMF, a novel and general framework for training NN-based multi-step time series forecasting models with known future information. It uses a masking technique that is flexible and can generate forecasts over different horizons. It improves existing methods by combining both recent history and known future information.
	\item The proposed framework has been applied to electricity demand forecasting in the energy sector, as well as flight departure forecasting in the transportation sector. MMMF shows consistently better results for different forecasting horizons on different models in these real-world datasets. %Experiments on two real-world datasets on electricity demand forecasting and flight departures forecasting show that MMMF outperforms existing time series forecasting tasks significantly with the exact same base NN model.
	\item MMMF performs inference at a similar speed and memory usage as existing methods with the same base model, making it suitable for implementation in real-world systems as an upgrade to existing approaches.
\end{itemize}

It should be emphasized that the goal of this paper is not to solve one multi-step forecasting problem with the best-tuned model, rather, to offer a new learning framework which could be applied to any NN model.
The comparisons with existing time series forecasting methods are by no means exhaustive, but they are fair because they use the same base model and hyperparameters, as well as representative because multiple NN models are evaluated.
It should also be noted that studying how future information (i.e., weather forecasts, future oil prices) is generated, or evaluating how good it is, are beyond the scope of this paper.
We simply focus on a general deep learning-based time series modeling framework that can incorporate future information in making multi-step forecasts when it is available.

\section{Related Work}~\label{sec:review}
\subsection{Multi-Step Time Series Forecasting}~\label{sec:review:msts}
Multi-step time series forecasting has been studied extensively.
Traditionally, regression models such as multiple linear regression and hidden Markov model~\citep{cheng2006multistep}, feed-forward neural networks~\citep{kline2004methods}, nearest neighbors~\citep{sorjamaa2007methodology}, decision trees~\citep{yang2009multi}, support vector regression~\citep{bao2014multi}, ensemble of varied length mixture models~\citep{ouyang2018multi} have been used for this problem.

Deep learning-based approaches can be categorized into recursive methods and direct methods. 
Recursive methods~\citep{li2019enhancing, salinas2020deepar, lim2020recurrent, suradhaniwar2021time} generate multi-step forecasts by recursively feeding 1-step forward predictions into future time steps.
This approach allows for easy generalization of 1-step forecasting models to multi-step forecasting, but the disadvantage is that the error would accumulate quickly because a forecast is based on all previous forecasts.
Direct methods~\citep{hauser2017probabilistic, fan2019multi, hauser2018neural, lim2021temporal, dabrowski2020forecastnet, masood2022multi} maps all available input directly to all forecasts, and sequence-to-sequence architecture is often used.
This approach requires a more complicated model structure and could be difficult to train.

These existing methods, however, do not use any known future information during training and are predicting the future with past information.
In MMMF, we directly incorporate available future information as well as past information.
In the extreme case that there is no future information, MMMF reduces to the aforementioned direct method. 

\subsection{Masked Time Series Models}~\label{sec:review:mask}
Following the success of BERT~\citep{devlin_bert_2019} and masked language models (MLMs) in learning language representations, there has been some work in applying the masking technique in time series modeling.
In time series generation, masked autoencoders~\citep{zha2022time} and masked generative adversarial networks (GANs)~\citep{srinivasan2022time} have been proposed.
In time series anomaly detection, masked models have been used in applications in spacecraft~\citep{meng2019spacecraft} and industrial data~\citep{fu2022mad}.
Masked time series models have also been used in reconstruction~\citep{haresamudram2020masked}, classification~\citep{liu2021gated}, representation learning for tabular data~\citep{padhi2021tabular}, etc.

Most of the existing work on masked time series chooses one specific model to mask and use that to compare against other models.
In this paper, we take a different approach and evaluate the entire idea of masking, and compare it with traditional regression tasks like next step prediction using the same underlying base model.

\subsection{Forecasting Electricity and Flight Demand}~\label{sec:review:application}
Forecasting electricity demand is essential for energy management, maintenance scheduling, energy trading, etc., and weather conditions play an important role in this problem~\citep{mirasgedis2006models}.
These forecasting processes can be grouped into four categories based on their horizons~\citep{hong2016probabilistic}: very short term, short term, mid term, and long term load forecasting, and there are thousands of papers on them.
Most relevant to this paper are deep learning-based short-to-mid-term forecasting models that require multi-step outputs.
These include CNN for short-term load forecasting~\citep{deng2019multi}, sequence-to-sequence LSTM for single house load forecasting~\citep{masood2022multi}, hybrid CNN and LSTM for short-term consumption forecasting~\citep{yan2018multi}, etc.

Forecasting demand for air travel is crucial for the commercial aviation industry.
As a real-world application of multi-step multivariate time series forecasting problem, it is very challenging due to multiple heterogeneous input variables~\citep{wang2022flight}.
For short-term demand forecasting, various traditional techniques including time series models, regression models, and hybrid models have been evaluated and compared~\citep{wickham1995evaluation}.
More recently, deep learning approaches including LSTM~\citep{wang2020deep} and Transformer\citep{wang2022flight} have been used in flight demand forecasting.

\section{Masked Multi-Step Multivariate Forecasting}\label{sec:mmmf}
\subsection{Problem Formulation}~\label{sec:mmmf:formulation}
Consider a multivariate time series forecasting problem: let $\mathbf{x_t}\in \mathbb{R}^n$ be a sample of predictor variables with dimension $n$ at time $t$ and the $j$-th dimension is denoted as $x^j_t$ (i.e., $\mathbf{x_t}=[x_t^1, x_t^2,...,x_t^n]$), $\mathbf{y_t}\in \mathbb{R}^m$ be a sample of forecast variables with dimension $m$ at time $t$ (i.e., $\mathbf{y_t}=[y_t^1, y_t^2,...,y_t^m]$), the task is to predict up to $(k+1)$ steps ($k \geq 0$) of forecast variables $\mathbf{y_t}, \mathbf{y_{t+1}}, ..., \mathbf{y_{t+k}}$ from past $T$-step information as well as some knowledge about the future predictor variables up to time $t+k$. 
It should be noted that this formulation can easily accommodate discrete-valued predictor variables by simply adding an embedding layer.
A distinct feature of this problem formulation compared to a standard time series forecasting problem is the need to incorporate future information into the predictions directly. 
For example, when forecasting electric demand for a particular region over the next month, the calendar variables (date, month, day of week, etc.) and weather forecasts are known, yet traditional forecasting formulations do not take full advantage of the future information.

\begin{figure*}[t]
	\centering
	\includegraphics[width=1.0\linewidth]{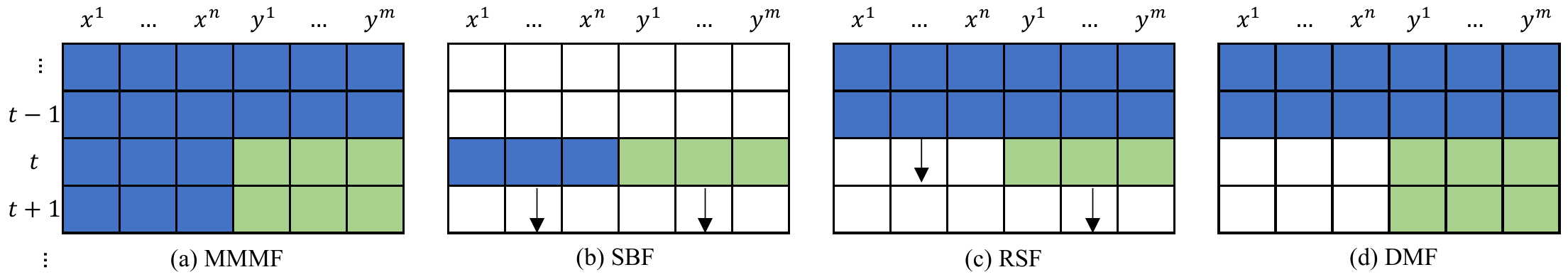}
	\caption{Comparison of multi-step forecasting formulations, variables in blue (darker shade) are used to forecast those in green (lighter shade). (a) Our proposed MMMF directly uses all available past and future information to predict all forecast variables. (b) SBF treats each future prediction separately, making forecasts with only predictor variables at that step and then advances to the next. (c) RSF makes a 1-step prediction on current forecast variables using past information, then advances the time window and makes predictions recursively. (d) DMF directly maps past information to multi-step future predictor variables.}
	\label{fig:formulation}
\end{figure*}

Formally, our proposed MMMF method directly models the following relationships:
\begin{equation}~\label{eqn:mmmf}
%	\begin{split}
		\hat{\mathbf{y_t}},...,\hat{\mathbf{y_{t+k}}} =  f(\mathbf{x_{t-1}},...,\mathbf{x_{t-T}},\mathbf{y_{t-1}},...,\mathbf{y_{t-T}},\mathbf{x_t},...,\mathbf{x_{t+k}})
%	\end{split}
\end{equation}
where $f$ is the function being modeled, $\hat{\mathbf{y_t}}$ are estimations of the ground truth $\mathbf{y_t}$ values, $\mathbf{x_{t-1}}, ... ,\mathbf{x_{t-T}}$ are the past predictor variables, $\mathbf{y_{t-1}}, ... ,\mathbf{y_{t-T}}$ are the past forecast variables, and $\mathbf{x_t}, ... ,\mathbf{x_{t+k}}$ are the future predictor variables.

Traditionally, there are three most common machine learning formulations for modeling such a multi-step multivariate forecasting problem.
\begin{enumerate}
	\item Sample-based forecasting (SBF) approach: this formulation treats each step as a distinct sample, and learns a function that maps the predictor variables to forecast variables directly without considering the temporal dependency, i.e., they model the following relationship:
	\begin{equation}~\label{eqn:sbf}
		\hat{\mathbf{y_{t}}} = f(\mathbf{x_{t}}), ... ,\hat{\mathbf{y_{t+k}}} = f(\mathbf{x_{t+k}})
	\end{equation}
	This non-time series direct mapping from input to output could use any traditional regression models, e.g., Linear Regression, fully connected neural networks, etc. However, it falls apart if there are no predictor variables but only forecast variables. Another disadvantage of this approach is that the temporal information is lost and recent history would not affect the forecasts.
	\item Recursive single-step forecasting (RSF) approach: this formulation is the standard next step prediction (NSP) task for a time series, where during training a one-step forward prediction model is learned, i.e., the loss is only calculated on the next step. That learned model is then being applied recursively during inference, i.e.:
	\begin{equation}~\label{eqn:rsf}
		\begin{split}
			\hat{\mathbf{y_{t}}} &= f(\mathbf{x_{t-1}},...,\mathbf{x_{t-T}},\mathbf{y_{t-1}},...,\mathbf{y_{t-T}}) \\ 
			\hat{\mathbf{y_{t+1}}} &= f(\mathbf{x_{t}},...,\mathbf{x_{t-T+1}},\hat{\mathbf{y_{t}}}, \mathbf{y_{t-1}},...,\mathbf{y_{t-T+1}}) \\
			&...
		\end{split}
	\end{equation}
	RSF does use all future information during training because the task is simply NSP.
	The major disadvantage of this formulation is that it makes predictions based on previous predictions, thus compounding errors will grow with the increasing number of steps. 
	\item Direct multi-step forecasting (DMF) approach: this formulation directly generates multiple outputs for all future steps of forecast variables in a time series, given past information, i.e.:
	\begin{equation}~\label{eqn:dmf}
		\hat{\mathbf{y_t}},...,\hat{\mathbf{y_{t+k}}} = f(\mathbf{x_{t-1}},...,\mathbf{x_{t-T}},\mathbf{y_{t-1}},...,\mathbf{y_{t-T}})
	\end{equation}
	DMF does not utilize the known future information and they simply map the past information to future predictions.
	Many base models for RSF, such as recurrent neural networks, could be reused for DMF. 
	The difference is how the outputs of those models are mapped, i.e., to 1-step future versus multi-step future forecast variables.
\end{enumerate}

These traditional techniques mainly suffer from two categories of issues. 
On one hand, SBF does not consider the temporal components and thus could perform poorly when the forecasting horizon is short. % is not sensitive to the most recent forecast variables' values. 
On the other hand, RSF and DMF do not utilize the knowledge of some future information for making forecasts. 
MMMF is proposed to take advantage of both information from the past and the future to make better forecasts.

It should be noted that the goal of this formulation is not to evaluate how good the future predictor variables are, but instead to develop a framework that could assimilate them regardless of how they are generated and use them to predict forecast variables. 
In real-world scenarios, some predictor variables are clearly defined and deterministic, like day of week, while others come with some uncertainty, like weather forecasts for the next week. 
The gap in existing formulations that MMMF addresses is that they cannot properly incorporate the known future information.
Therefore, MMMF is a more general time series modeling framework than traditional regression models, and if there is no known future information, MMMF reduces to an autoregressive-like masked DMF model.

\subsection{Proposed Solution}~\label{sec:mmmf:solution}
To solve this multi-step multivariate forecasting problem, our proposed MMMF as shown in Figure~\ref{fig:formulation}~(a) directly uses a masked time series model (MTSM) approach. 
Inspired by~\citep{fu2022mad} where MTSMs are used for time series anomaly detection and have shown superior performance to traditional regression models, we propose the method for MMMF training as given in Algorithm~\ref{algo:mmmf-train}.
The time series model $f_\theta$ in this algorithm (or the base model for MMMF) can be any neural network model that generates a sequence of outputs, such as Long Short-Term Memory (LSTM) network~\citep{hochreiter_long_1997}, Temporal Convolutional Network (TCN)~\citep{bai_empirical_2018} Transformer~\citep{vaswani_attention_2017}, etc.
Therefore, MMMF is not limited to one model but is a general learning task for all time series NN models.

\begin{algorithm}[!ht]
	\KwIn{Time series model $f_{\mathbf{\theta}}$ with a set of trainable parameters $\mathbf{\theta}$, maximum forecasting horizon $k$, maximum history length $T$, loss function $\ell$}
	\KwData{Time series dataset $S=\{\mathbf{z_i}\}=\{(\mathbf{x_i}, \mathbf{y_i})\}$, where $i$ represents the $i$-th time step, $\mathbf{x_i}$ are the predictor variables, $\mathbf{y_i}$ are the forecast variables}
	Preprocessing dataset with a sliding window of length $(T+k+1)$ to $\{\mathbf{z_{t-T}},...,\mathbf{z_{t-1}},\mathbf{z_{t}},\mathbf{z_{t+1}},...,\mathbf{z_{t+k}}\}$ sequences, where $\mathbf{z_{t}}$ is the sample at current step\;
	Initializing model parameters $\mathbf{\theta}$\;
	\While{not at end of training epochs}{
		\While{not at the end of all mini-batches}{
			Randomly choose a batch of sequences\;
			Randomly choose an integer mask length $l_m$ for this current batch, $0<l_m\leq k+1$\;
			\For{each sequence in the mini-batch}{
				Mask last $l_m$ steps of forecast variables $\mathbf{y}$\;
			}
			Feed masked sequences to model $f_{\mathbf{\theta}}$, generate estimations $\hat{\mathbf{y}}$\ using information of predictor variables from both the past and future $\mathbf{x_{t-T}}, ..., \mathbf{x_{t+k}}$, and unmasked forecast variables $\mathbf{y_{t-T}}, ..., \mathbf{y_{t+k-l_m}}$\;
			Calculate loss only on the masked outputs for future predictions $\sum_{i=k-l_m}^{i=k}\ell(\mathbf{y_i},\hat{\mathbf{y_i}})$\;
			Backpropagation, update model parameters $\mathbf{\theta}$\;
		}
	}
	\KwOut{Trained model $f_{\mathbf{\theta}}$}
	\caption{MMMF Training}
	\label{algo:mmmf-train}
\end{algorithm}

The key step for Algorithm~\ref{algo:mmmf-train} is Step 9, i.e., random masking of the last $l_m$-step of forecast variables in a sequence.
Because $l_m$ is chosen randomly for each mini-batch of data, this essentially creates many forecasting sub-tasks where at each iteration the base model $f_\theta$ is trying to forecast a different length outputs.
In one extreme, when $l_m=1$, MMMF reduces to a similar formulation as RSF in Figure~\ref{fig:formulation}~(c) with the exception that the information of predictor variables at time step $t$ is also used; in the other extreme, when $l_m=k+1$, MMMF reduces to a similar formulation as DMF in Figure~\ref{fig:formulation}~(d) with the exception that the information of predictor variables from time step $t$ to $t+k$ is also used.
From this perspective, it can be seen that the learning task of MMMF is more comprehensive than the traditional time series regression tasks, as well as the non-time series SBF regression task in Figure~\ref{fig:formulation}~(b).

Different from Masked Language Models (MLMs) such as BERT~\citep{devlin_bert_2019} where the tokens are discrete, time series data are often continuous.
Therefore, MMMF replaces all masked variables with random values within the ranges of those variables.
Different from autoencoders, MMMF calculates the loss on only the masked outputs, which is similar to other masked approaches such as BERT, instead of the full reconstruction loss.

Another major advantage of the MMMF learning task, which uses all the rest information to forecast the variable-length masked variables, is that once trained a base neural network model can generate forecasts for any forecast length $l_f$ for $0<l_f \leq k+1$, by simply masking the last $l_f$ steps of the desired forecast variables.
The MMMF inference is given in Algorithm~\ref{algo:mmmf-inference}.

\begin{algorithm}[!ht]
	\KwIn{MMMF-trained time series model $f_{\mathbf{\theta}}$, forecast horizon $l_f$ where $0<l_f\leq k+1$}
	\KwData{A sequence of length $(T+k+1)$, with all predictor variables $\mathbf{x}$ known, and last $l_f$-step forecast variables $\mathbf{y}$ unknown}
	Fill last $l_f$-step forecast variables $\mathbf{y}$ with mask\;
	Provide masked sequences to trained model $f_{\mathbf{\theta}}$\;
	Generate forecasts $\hat{\mathbf{y}}$ for the last $l_f$ steps\;
	\KwOut{Multi-step multivariate forecast for forecast variables $\mathbf{y}$ of length $l_f$}
	\caption{MMMF Inference}
	\label{algo:mmmf-inference}
\end{algorithm}

Because an MMMF-trained model has learned to generate different lengths of forecasts during its training process, it is very flexible during inference and could generate any length of forecasts from $1$ to the maximum forecast horizon $k$.
Fundamentally, the self-supervised learning approach should learn a representation of the data by being able to fill in the blanks when some forecast variables are masked.
This leads to the flexibility of MMMF-trained models during inference: they are not restricted to making fixed-length forecasts.
This could potentially be useful in some real-world applications, e.g., when an electricity load demand forecast model is trained, it needs to be able to make both short-term forecasts for unit commitment and mid-term forecasts for fuel planning and maintenance planning.
Instead of having multiple models for each application, an MMMF-trained model could do all of them.

Furthermore, since masking requires very little additional computational time, MMMF-trained models could generate forecasts at a similar speed and memory usage as RSF and DMF approaches if they are using the same base model.
Given the more complicated learning task, it may take longer for MMMF models to converge during training, but in practice, the inference time is more important for real-world applications.
%Experiment results in Section~\ref{sec:exp} will validate this point.

\section{Experimental Results}~\label{sec:exp}
%sensitivity/robustness study on weather is probably better suited for another paper focused on weather-load forecasting
In this section, we apply Masked Multi-Step Multivariate Forecasting (MMMF) to two datasets: ISO New England electricity demand forecasting dataset and regional flight departures forecasting dataset, and compare them with the other existing forecasting formulations as shown in Figure~\ref{fig:formulation}.
Both of these real-world datasets have some known information about the future that enables comparison among different approaches.
For a fair comparison, MMMF, RSF, and DMF all use the same base models for forecasting, while SBF is not a time series approach so the base model is different.

\subsection{Mid-Term Electricity Demand Forecasting}~\label{sec:exp:electricity}
In this subsection, we use a real-world electricity energy demand forecasting dataset to compare our proposed MMMF method against existing methods.
The task is to forecast the electricity demand for up to 60-day ahead, given the calendar variables and weather information.

\subsubsection{Dataset and Base Models}~\label{sec:exp:electricity:dataset}
For this daily peak electricity demand forecasting problem, we use the ISO New England zonal dataset\footnote{Raw data files can be downloaded at \url{https://www.iso-ne.com/isoexpress/web/reports/load-and-demand/-/tree/zone-info}} from the year 2011 to 2021. 
The data from 2011 to 2020 are used for training, and the data from 2021 are used for testing.

The entire ISO New England is divided into 8 different zones: Connecticut (CT), Maine (ME), Northeast Massachusetts and Boston (NEMA), Hew Hampshire (NH), Rhode Island (RI), Southeast Massachusetts (SEMA), Vermont (VT) and West/Central Massachusetts (WCMA). 
For unit commitment, fuel planning, or maintenance planning, daily peak electricity demand forecasting from 1-day ahead up to 60-day ahead is often of interest~\citep{hong2016probabilistic}.
Therefore, the original hourly dataset is downsampled to daily by taking the maximum values of that day.
Predictor variables include month, date, day of week, dry bulb temperature, and dew point temperatures for each zone.
Forecast variables are the electricity demand for each zone, so this is a multivariate problem with 8 variables to forecast.

For this multi-step multivariate forecasting problem, MMMF uses a total window length of 90 and the maximum forecast length is 60, i.e., a 30-step history is always known.
RSF also uses an input length of 30 and output length of 1 during training but is being applied recursively for 60 steps with future information during testing to generate the multi-step forecasts.
DMF directly maps a sequence of input length 60 to an output length of 60 for this task.
For these three time series modeling frameworks, they use the exact same underlying base model and hyperparameters for all following experiments to perform fair comparisons, which include:
\begin{itemize}
	\item \textbf{LSTM}~\citep{hochreiter_long_1997}: $2$ hidden layers, each with hidden dimension of $50$.
	\item \textbf{TCN}~\citep{bai_empirical_2018}: $2$ hidden layers with channel size $50$ each, convolutional kernel size $3$ and stride $1$, dilation factor is $2^i$ where $i$ is the $i$-th layer, and dropout rate $0.2$.
	\item \textbf{Transformer}~\citep{vaswani_attention_2017}: model dimension $128$, feed-forward dimension $512$, number of heads $8$, number of encoder  layers $2$, and dropout rate $0.1$. Only an encoder is used.
\end{itemize}
For SBF, a $2$-layer fully connected neural network with ReLU activation is used.

For all experiments, we trained the models on a Nvidia V100 GPU using an Adam optimizer~\citep{kingma_adam_2015} with $\beta_1=0.9$, $\beta_2=0.999$, $\epsilon=10^{-8}$ and a learning rate of $0.001$.
Model batch size is set to be $1000$ and the number of epochs is $1000$. 
The metric for calculating loss during training is mean squared error (MSE) loss, but we report the mean absolute percentage error (MAPE) for all test results in this subsection following common practices in the field of electricity demand forecasting.
For training sets, 80\% of the data were used for training and 20\% for validation.
For categorical variables, an embedding size of $5$ is used.
All implementations are written in PyTorch~\citep{NEURIPS2019_9015}.

\subsubsection{Multi-Step Multivariate Forecasting}~\label{sec:exp:electricity:ms}
\begin{figure*}[t]
	\centering
	\begin{subfigure}[b]{0.33\textwidth}
		\centering
		\includegraphics[width=\textwidth]{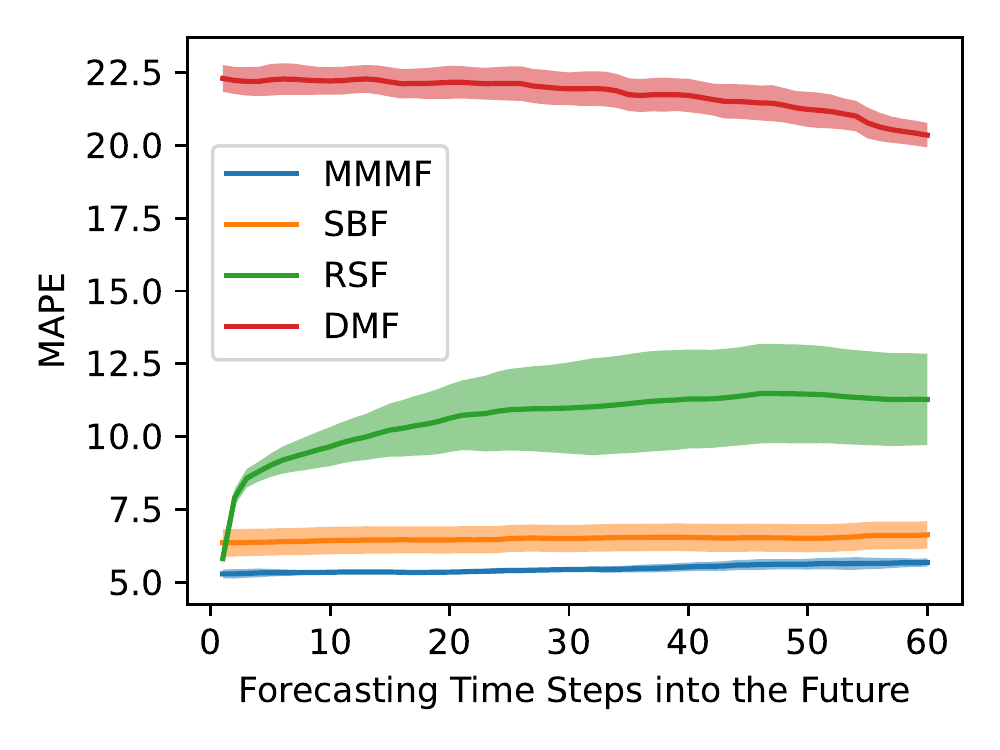}
		\caption{LSTM}
		\label{fig:elec:lstm_60}
	\end{subfigure}
	\begin{subfigure}[b]{0.33\textwidth}
		\centering
		\includegraphics[width=\textwidth]{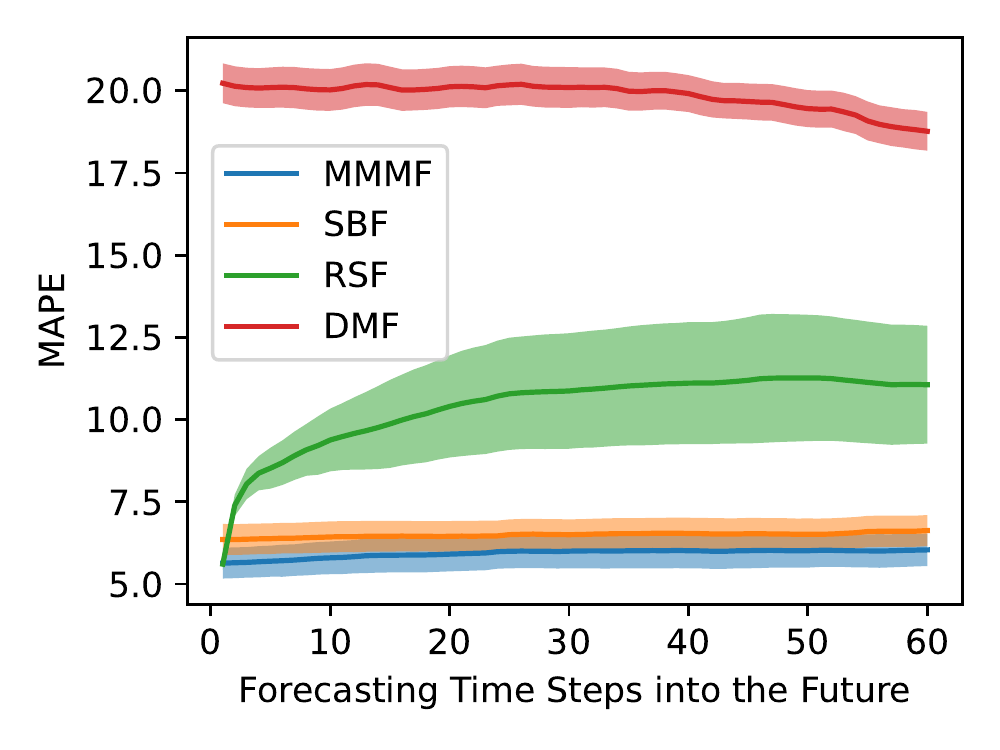}
		\caption{TCN}
		\label{fig:elec:tcn_60}
	\end{subfigure}
	\begin{subfigure}[b]{0.33\textwidth}
		\centering
		\includegraphics[width=\textwidth]{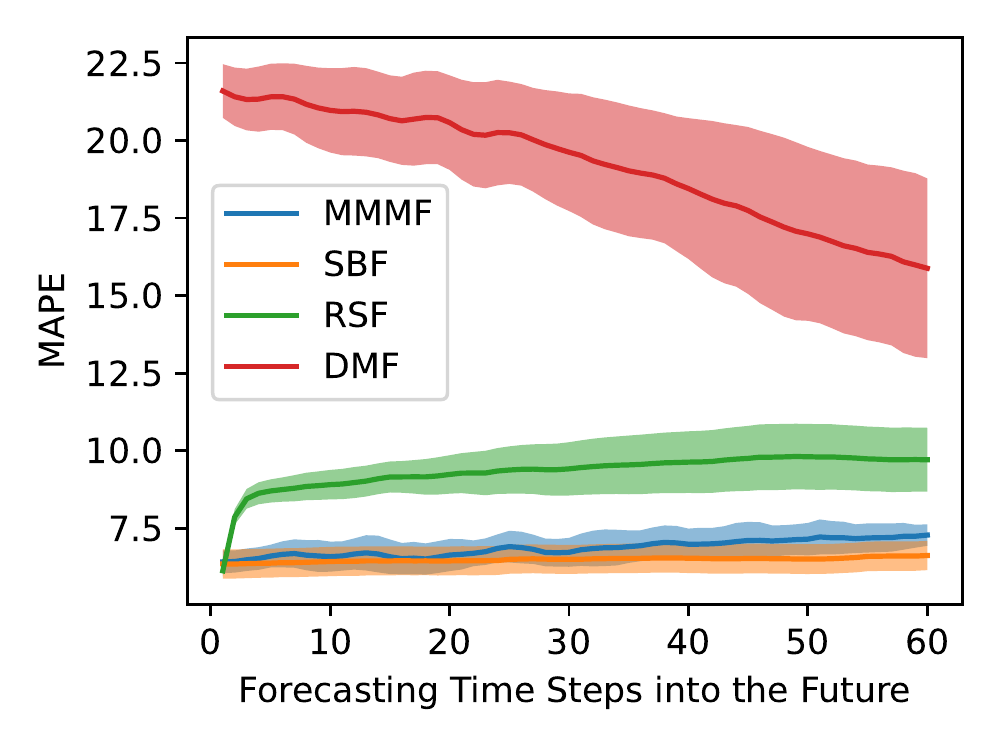}
		\caption{Transformer}
		\label{fig:elec:transformer_60}
	\end{subfigure}
	\caption{60-step daily peak electricity demand forecasting. The mean and standard deviation of MAPE for different methods of all 8 zones are shown. Base models for MMMF, RSF, and DMF are the same in each figure.}
	\label{fig:elec}
\end{figure*}

The results for 60-day ahead zonal level electricity demand forecasting is shown in Figure~\ref{fig:elec} and grouped by different base models, i.e., MMMF, RSF, and DMF use the same base model in each figure.
SBF uses the same fully connected NN model across the three figures and provides a baseline for comparison.
Models are trained with different random seeds and the mean and standard deviation of 5 runs for each training method are reported in this figure.
Note that the MAPE in these plots is the average of all 8 different zones.

Figure~\ref{fig:elec} shows that for the multi-step forecasting problem with known future information, MMMF significantly outperforms RSF and DMF with exactly the same base model and hyperparameters.
Because RSF is trained on 1-step forward predictions, it performs relatively well at the beginning, but the performance degrades significantly as the number of steps increases and predictions are made on previous predictions.
That is to say, the recursive approach as given in Equation~\ref{eqn:rsf} is not suitable for this multi-step forecasting problem.
The performance of DMF is the worst because it needs to directly map the historical data to all 60-step forecasts without properly using the known future information.
Since SBF uses a completely different model, the performance difference largely depends on this specific dataset and how the other base models perform.
In this electricity demand forecasting problem where the weather and calendar information plays an important role, SBF provides a strong baseline.

\begin{table}[ht]
	\centering
	\caption{Mid-term electricity demand forecasting results. Average MAPE for the 60-day ahead forecasting is reported for different methods (lower is better, best among same base model is in bold). The inference time (in seconds) for the entire dataset (100 runs averaged) is also reported.}
	%	\resizebox{0.88\linewidth}{!}{
		\begin{tabular}{c|c|c|c}
			\toprule
			Base Model & Method & MAPE & Inference Time \\
			\midrule
			\multirow{3}{*}{LSTM} & MMMF & \textbf{5.46} & 0.0358$\pm$0.0018s \\ 
			& RSF & 10.62 & 0.0342$\pm$0.0017s \\
			& DMF & 21.75 & 0.0362$\pm$0.0085s \\
			\midrule
			\multirow{3}{*}{TCN} & MMMF & \textbf{5.93} & 0.0258$\pm$0.0063s \\
			& RSF & 10.39 & 0.0234$\pm$0.0016s \\
			& DMF & 19.84 & 0.0214$\pm$0.0020s \\
			\midrule
			%			\multirow{2}{*}{\shortstack{Transformer-\\Encoder}}
			\multirow{3}{*}{Transformer} & MMMF & \textbf{6.86} & 0.0262$\pm$0.0013s \\ 
			& RSF & 9.32 & 0.0561$\pm$0.0187s \\
			& DMF & 19.19 & 0.0317$\pm$0.0198s \\
			\midrule
			NN & SBF & 6.49 & 0.0068$\pm$0.0171s \\
			\bottomrule
		\end{tabular}
		%	}
	\label{table:isone}
\end{table}

Furthermore, we summarize both the results for Figure~\ref{fig:elec} and the model inference time in Table~\ref{table:isone}. 
The inference time only counts the GPU time, after dataset loading and preprocessing, and is averaged over 100 inference runs on the same hardware.
It can be seen that once a base model is trained with MMMF, its inference speed is similar to that trained with traditional RSF or DMF formulations because the only additional step in MMMF is random masking, which is not computationally expensive.
In the case of Transformer base model, since it is naturally suited for a masked forecasting framework, the inference speed is faster than using it iteratively to generate 1-step ahead forecasts for 60 times.
Since the base model is the same, they all have the same memory usage, so if a system can run an RSF-trained model for inference, it can run MMMF-trained models, too.
It shows that for many deployed real-world forecasting systems which were trained using RSF or DMF, switching to MMMF would lead to a significant performance increase without sacrificing speed or requiring any hardware upgrade.

\subsubsection{One-Step Multivariate Forecasting}~\label{sec:exp:electricity:1s}
Here we zoom in on the 1-step forecasting results in Figure~\ref{fig:elec}.
This reduces the problem to the standard next step prediction task in time series forecasting and is of significant importance to many short-term problems in the real world.

\begin{figure}[t]
	\centering
	\includegraphics[width=\linewidth]{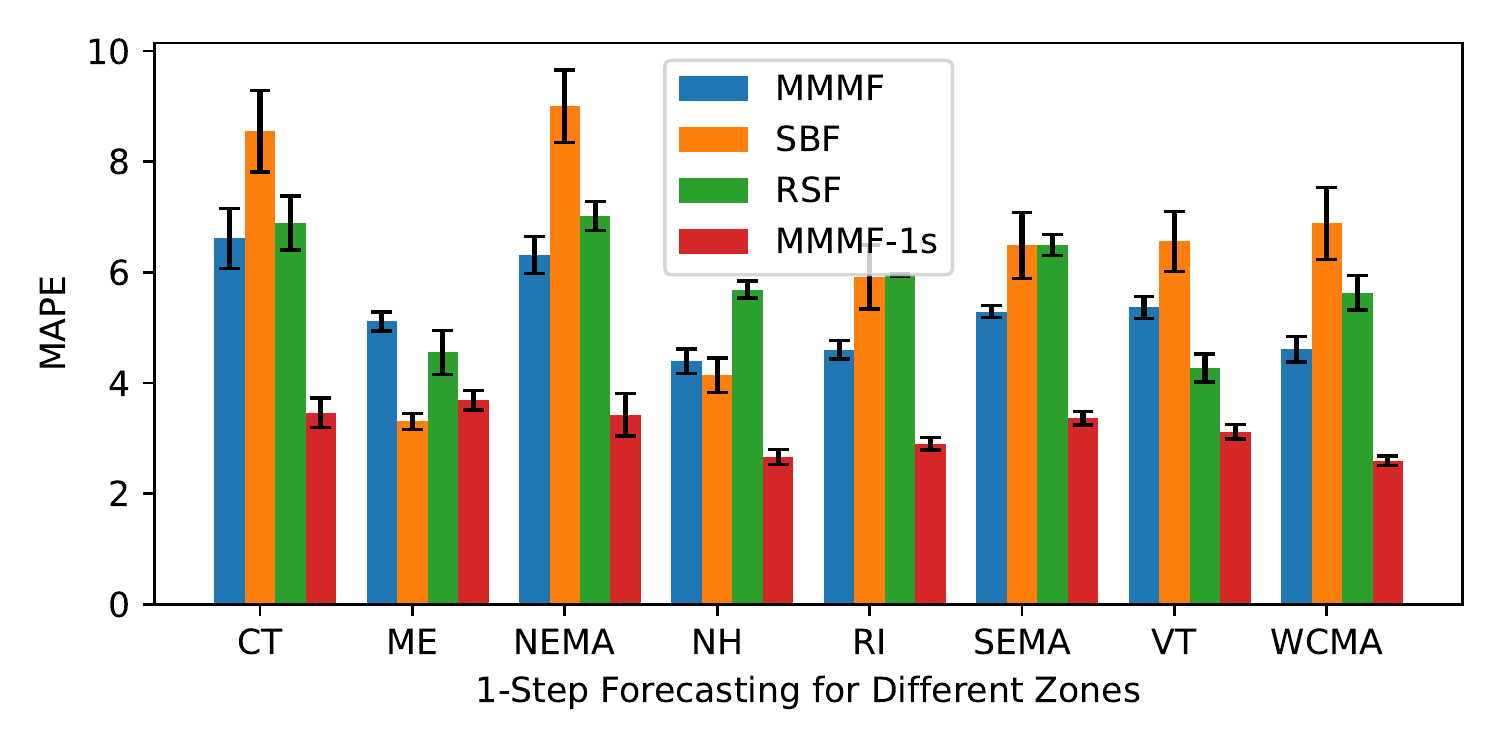}
	\caption{Comparison of 1-step forecasting results for different zones. The mean and standard deviation of MAPE is shown for each method. MMMF and RSF are the same LSTM base model as in Figure~\ref{fig:elec:lstm_60}, and MMMF-1s use the same LSTM model with MMMF training technique and a maximum mask length of 1.}
	\label{fig:lstm_1step}
\end{figure}

We compare the results of MMMF, RSF (with LSTM base model), and SBF for one-step forecasting from the previous subsection in Figure~\ref{fig:lstm_1step}. 
The MAPE for different zones is plotted separately.
Note that MMMF is trained for forecasting up to 60 steps, but is capable of generating forecasts for 1 step, whereas RSF is directly trained on this 1-step forecasting.
To facilitate fair comparisons, we trained an additional MMMF model on 1-step only, i.e., the maximum mask length is 1 instead of 60 during training and labeled this model as `MMMF-1s' in Figure~\ref{fig:lstm_1step}. 
We did not retrain a DMF model for 1-step only forecasting, because in that case DMF is reduced to exactly RSF as seen in Figure~\ref{fig:formulation}.

From these 1-step forecasting results in Figure~\ref{fig:lstm_1step}, we can see that SBF performs the worst because it does not take into account the recent history when making a prediction.
MMMF trained on up to 60-step ahead forecasting performs slightly better than RSF, and MMMF-1s which is trained specifically on this 1-step forecasting task with masks, perform the best overall.
This result shows the importance of incorporating the known future information, even for short-term predictions: the difference between MMMF-1s and the traditional next step prediction in time series (RSF) is that MMMF-1s adds the future information for that step.
In the context of this electricity demand forecasting problem, this means that MMMF uses tomorrow's weather directly as well as recent history, thus making it better than only relying on historical data (RSF) or only relying on tomorrow's weather but no recent data (SBF).

\subsection{Regional Flight Departures Forecasting}~\label{sec:exp:flight}
In this subsection, to illustrate the general applicability of the MMMF framework, we demonstrate it on a real-world forecasting problem in the commercial aviation industry.
Forecasting demands for air travel is a challenging problem but critical for understanding aviation market behavior.
We evaluate two dataset setups (with and without a predictor variable on jet fuel price) on different forecasting formulations.
%Traditionally, regional air traffic can be modeled as a function of economic growth, aviation infrastructure growth and local regulation governance, weather conditions, seasonality, holidays, and operator dynamics as well as other factors. Non-traditional interruptions, such as oil and financial crises, terrorism, wars, and pandemics, can also have significant impacts. Models of aviation market trends measured through departures, therefore, allow the business to be prepared for various types of market interruptions that may occur in the future.

\subsubsection{Dataset Description}~\label{sec:exp:flight:dataset}
For this real-world dataset, the number of daily passenger aircraft departures for each of seven regions (North America, Latin America, Europe, CIS, APAC, China, MEA) across two aircraft classes (Narrowbody/Widebody) with GE and CFM-powered engines is collected from FlightAware.
Historical data on each region's daily departures from January 2019 through February 2022 are used as the training set, and the goal is to forecast the daily departures from 1 March 2022 through 30 April 2022 (61 days, 14 forecast variables for each day).
Similar to the previous subsection on electricity demand forecasting, we extract the calendar variables including month, date, day of week as predictor variables.
As an additional predictor variable, the monthly historical and future jet fuel price is provided by the short-term energy outlook from the U.S. Energy Information Administration (EIA)\footnote{The data on energy prices can be downloaded at \url{https://www.eia.gov/outlooks/steo/xls/STEO_m.xlsx}}.
In this dataset, we used the short-term energy outlook published by EIA at the end of February 2022, before the actual forecasting period.
This multi-step multivariate forecasting problem is particularly challenging because the daily regional flight departures are not only impacted by the predictor variables included in this dataset, but also by external factors such as pandemics, wars, oil and financial crises, etc.

\subsubsection{Two-Month Ahead Forecasting Results}~\label{sec:exp:flight:2m}
Using exactly the same base model setups and training hyperparameters, except that the two-month ahead forecasting spanning March to April 2020 includes 61 days instead of 60 days, we report the MSE for the forecasting period across all regions in Table~\ref{table:departure}. Unlike in Section~\ref{sec:exp:electricity} where the data across different regions are on the same order of magnitude, since the daily aircraft departures across different regions and aircraft classes vary from several dozens to thousands, MAPE can be slightly misleading so it is not reported for this dataset.

\begin{table}[ht]
	\centering
	\caption{Regional flight departure forecasting results with different training methods and dataset setups (without and with jet fuel price as a predictor variable). The lowest MSE for each base model is in bold.}
	%	\resizebox{0.88\linewidth}{!}{
		\begin{tabular}{c|c|c|c}
			\toprule
			Base Model & Method & w/o Fuel Price & w/ Fuel Price \\
			\midrule
			\multirow{3}{*}{LSTM} & MMMF & 850.38 & \textbf{734.74} \\ 
			& RSF & 1248.90 & 817.51 \\
			& DMF & 1367.47 & 1096.15 \\
			\midrule
			\multirow{3}{*}{TCN} & MMMF & 1155.36 & \textbf{688.19} \\
			& RSF & 1224.11 & 950.40 \\ 
			& DMF & 1283.83 & 983.12 \\
			\midrule
			%			\multirow{2}{*}{\shortstack{Transformer-\\Encoder}}
			\multirow{3}{*}{Transformer} & MMMF & 1206.00 & \textbf{828.04} \\ 
			& RSF & 1297.72 & 862.34 \\
			& DMF & 2080.69 & 1190.82 \\
			\bottomrule
		\end{tabular}
		%	}
	\label{table:departure}
\end{table}

The results show that with the same base model and dataset setup, MMMF again outperforms the traditional RSF and DMF formulations.
Furthermore, the MSE with the additional monthly jet fuel price as a predictor variable is lower than that without, showing the importance of incorporating known future information in real-world forecasting problems, even when the future information is not perfect and lower-resolution (in this case, jet fuel price projections are monthly but the departure forecasts are daily).
MMMF as designed can directly utilize known future information for multi-step multivariate forecasts, thus making it the ideal framework for these time series forecasting problems.

\section{Conclusion}~\label{sec:conclusion}
In this paper, we proposed and experimentally validated Masked Multi-Step Multivariate Forecasting (MMMF), a new self-supervised learning framework for multi-step time series forecasting with known future information.
The main contribution of this paper is not to find a new model and a set of hyperparameters for a particular problem, but to show that MMMF as a general training task can outperform existing time series forecasting approaches, including recursive methods and direct methods while using the same base model.
Once trained with MMMF, a time series model can generate any length forecasts below the maximum forecast length during training, and the inference speed, as well as the memory usage, are similar to those of traditional methods.
This makes MMMF an ideal upgrade to any existing deep learning-based multi-step time series forecasting models, and potentially has significant impacts on many real-world forecasting applications where some future information is available.

\bibliographystyle{abbrvnat}
\bibliography{mmmf}  %%% Uncomment this line and comment out the ``thebibliography'' section below to use the external .bib file (using bibtex) .





% \end{thebibliography}

\end{document}